\documentclass[10pt,twocolumn,letterpaper]{article}

\usepackage[pagenumbers]{iccv} 

\definecolor{iccvblue}{rgb}{0.21,0.49,0.74}
\usepackage[breaklinks,colorlinks,allcolors=iccvblue]{hyperref}
\usepackage[table]{xcolor}
\definecolor{lightgray}{gray}{0.9}
\definecolor{deepgreen}{rgb}{0,0.5,0}

\usepackage[super]{nth}

\def\paperID{4082} 
\def\confName{ICCV}
\def\confYear{2025}

\title{QK-Edit: Revisiting Attention-based Injection
in MM-DiT \\ for Image and Video Editing}

\author{
Tiancheng Shen$^{1,2,3,4}$ \quad
Zilong Huang$^{2}$ \quad
Xiangtai Li$^{2}$ \quad
Zhijie Lin$^{2}$ \quad
Jiyang Liu$^{3}$ \\
Yitong Wang$^{3}$ \quad
Jiashi Feng$^{2}$ \quad
Ming-Hsuan Yang$^{1}$ \quad
Jun Hao Liew$^{2}$ \\[0.4em]
$^{1}$\,UC Merced \qquad
$^{2}$\,ByteDance Seed \qquad
$^{3}$\,ByteDance Intelligent Creation \qquad
$^{4}$\,CUHK
}

\begin{document}

\twocolumn[{%
\renewcommand\twocolumn[1][]{#1}%
\maketitle
\begin{center}
    \centering
    \vspace{-1em}
    \includegraphics[width=\textwidth]{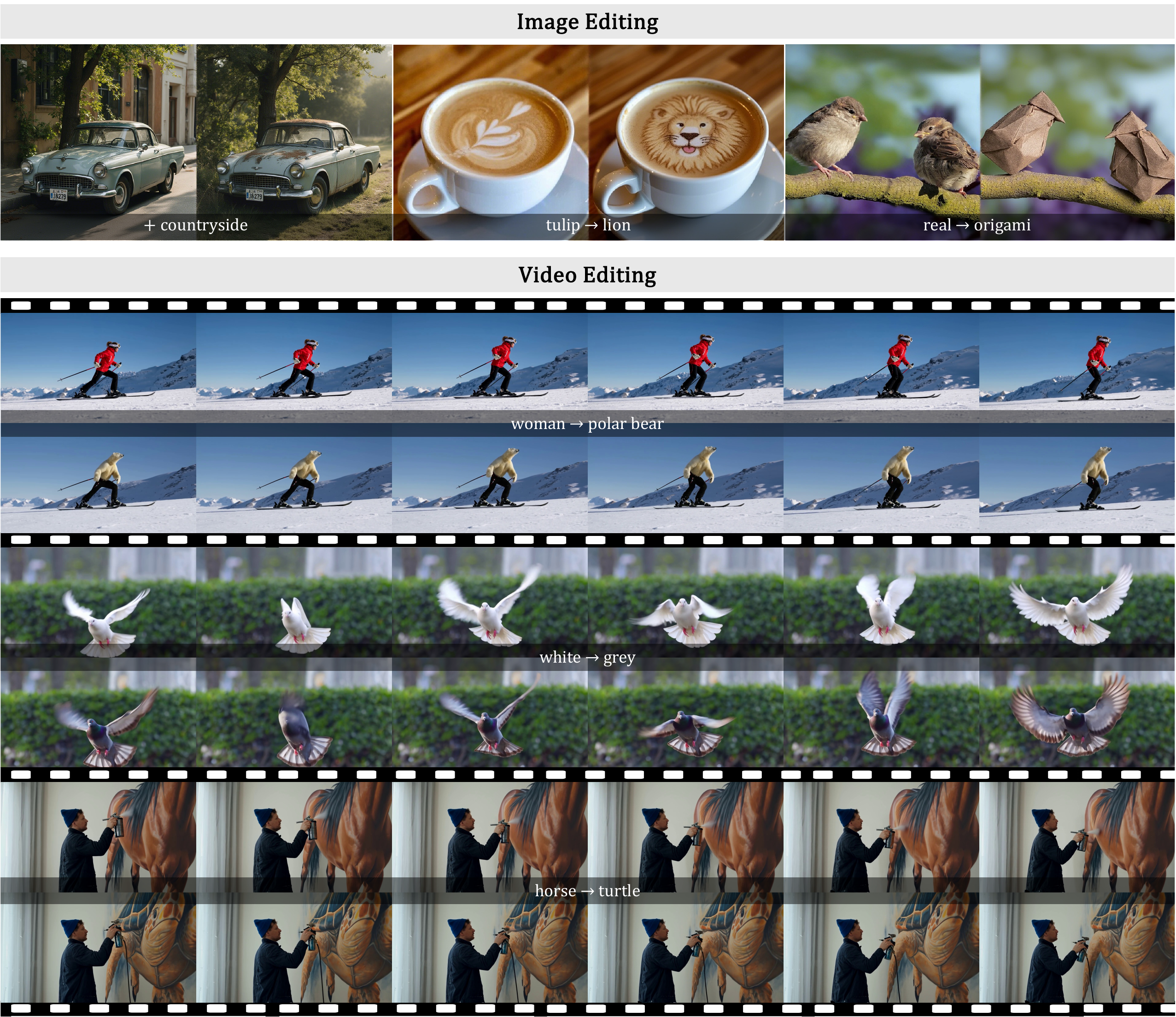}
    \captionof{figure}{\textbf{QK-Edit: A training-free, text-guided framework for high-fidelity image and video editing.} Our method leverages strategic query-key manipulation within the vision components of Multimoda Self-Attention~(MM-SA) in MM-DiTs to seamlessly perform diverse edits~(\eg, object addition, object removal, object replacement, changing background, changing material, changing color, and style transformation \etc), while preserving structural consistency and visual details. Please refer to the supplementary material for more video results at higher resolutions~(1K), different aspect ratios~(landscape and portrait formats) and longer lengths~(129 frames).}
    \label{fig:teaser}
\end{center}%
}]

\maketitle

\begin{abstract}

Multimodal Diffusion Transformers~(MM-DiTs) have recently emerged as a powerful framework for unified text-vision synthesis, surpassing traditional U-Net architectures in generative tasks. One key innovation lies in its Multimodal Self-Attention~(MM-SA) interaction where image and text tokens are concatenated and processed via self-attention.
However, this mechanism poses significant challenges for editing, rendering conventional U-Net-based attention manipulation methods ineffective. To address this limitation, we propose QK-Edit, a training-free framework that exploits the unique attention dynamics of MM-DiTs for precise text-guided image and video editing. By introducing a novel query-key manipulation strategy, our method isolates and adjusts critical attention components to achieve an optimal balance between prompt fidelity and structural consistency. This enables seamless edits across various tasks, including object addition, object removal, object replacement, changing background, changing material, changing color, and style transformation. 
Notably, it can be easily implemented with feature replacement in inference.
QK-Edit demonstrates superior editing performance on state-of-the-art models, such as FLUX and HunyuanVideo, effectively bridging the gap between generative power and editable flexibility in MM-DiTs, and paving the way for scalable multimodal content creation. 

\end{abstract}
\section{Introduction}
\label{sec:intro}

\begin{figure}[t]
    \centering
    \includegraphics[width=\columnwidth]{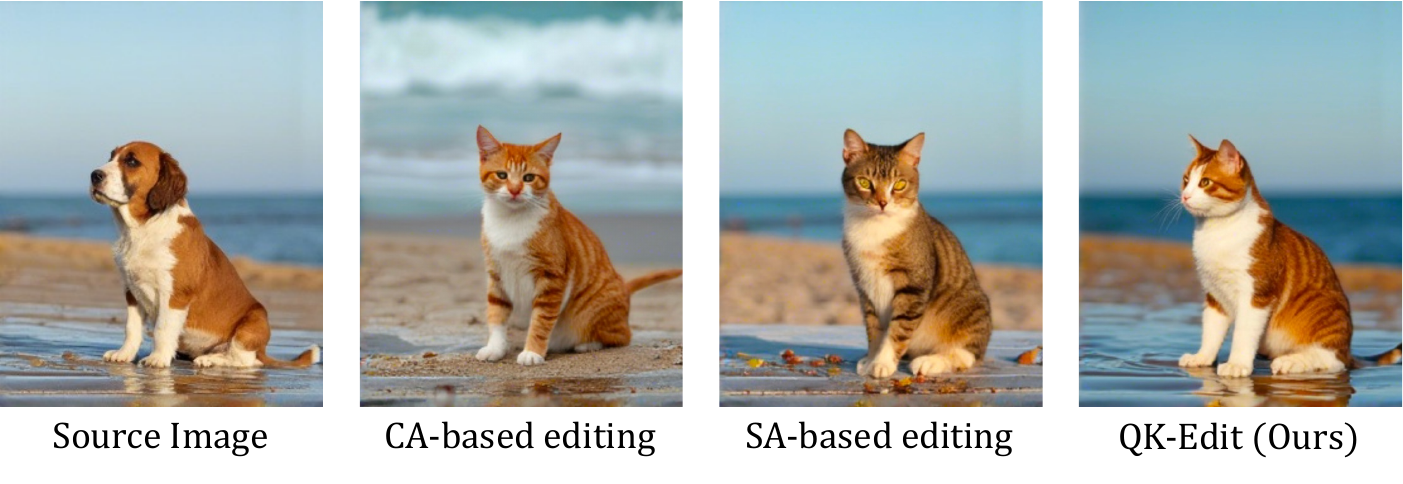}
    \vspace{-8mm}
    \caption{Directly applying CA-based and SA-based editing techniques designed for U-Net diffusion models to MM-DiTs can lead to undesirable results. 
    }
    \label{fig:Sec3.1_DirectUseFailure}
\end{figure}

The emergence of Transformer-based diffusion models~\cite{peebles2023scalable}, particularly Multimodal Diffusion Transformers (MM-DiTs)\cite{flux, sd3, kong2024hunyuanvideo, yang2024cogvideox}, has redefined the landscape of generative models for image and video synthesis.
Unlike U-Net architectures\cite{ronneberger2015u}, which separate text-vision Cross-Attention (CA) and vision-vision Self-Attention (SA), MM-DiTs leverage a Multimodal Self-Attention (MM-SA) mechanism to process multimodal inputs in a unified embedding space.
MM-SA concatenates text and vision embeddings and computes self-attention on the joint embedding\footnote{In this paper, “vision” refers to video or image data.}. This integration enables nuanced and coherent outputs, making MM-DiTs a cornerstone in models such as FLUX~\cite{flux}, Stable Diffusion 3.0~\cite{sd3}, HunyuanVideo~\cite{kong2024hunyuanvideo}, and CogVideo-X~\cite{yang2024cogvideox}. However, MM-SA introduces complexities that challenge existing editing paradigms based on attention manipulation~\cite{hertzprompt, Tumanyan_2023_CVPR}.
Specifically, due to the joint attention mechanism, textual and visual interactions become deeply intertwined within a single attention matrix, complicating the independent manipulation of textual and visual information. As a result, such editing methods for U-Net models become ineffective in Fig.~\ref{fig:Sec3.1_DirectUseFailure}.

These complexities arise because such previous editing methods rely heavily on the modular nature of U-Net architectures~\cite{rombach2022high}, where cross-attention facilitates precise text-guided modifications, and self-attention ensures spatial consistency. In contrast, MM-DiTs integrate these attention mechanisms into a unified structure, making it challenging to independently control text-vision and vision-vision interactions without one adversely affecting the other. 
A naive solution is to decompose the MM-SA map into four quadrants, each representing a specific interaction between text and vision embeddings and directly apply previous CA-based or SA-based editing techniques such as P2P~\cite{hertzprompt} or PnP~\cite{Tumanyan_2023_CVPR} to the respective attention maps. However, this often leads to undesirable outputs, as illustrated in Fig.~\ref{fig:Sec3.1_DirectUseFailure}.

Through systematic analysis, we observe: (1) text-to-vision attention has negligible impact on the final generation, (2) vision-to-text attention relates vision region to the text token, and finally (3) vision-vision attention defines both spatial and temporal consistency within visual embeddings. 
As a result, simultaneous replacement of vision-to-text and vision-to-vision attention is necessary to enable precise control over spatial and temporal properties while adhering to the target prompt. Building upon these insights, we propose QK-Edit, a training-free editing framework that manipulates the attention quadrants of MM-DiTs to achieve high-fidelity edits. In practice, we leverage the fact that text-to-vision attention has minimal impact on the generation and replace the vision component in query and key embeddings of the target image or video with that of the source image or video during the denoising process.
In practice, our proposed strategy is straightforward to implement, requiring only minimal code modifications.


Our QK-Edit can be seamlessly extended to image and video domains, demonstrating its versatility in various generative tasks.
Extensive experiments validate the effectiveness of QK-Edit on state-of-the-art models such as FLUX~\cite{flux} and HunyuanVideo~\cite{kong2024hunyuanvideo}, achieving superior results in tasks ranging from object-level modifications to global scene transformations. 
Notably, our method enables high-resolution, long-frame video editing with spatio-temporal consistency, overcoming the limitations of previous editing approaches. 
Furthermore, the training-free nature of QK-Edit ensures computational efficiency and broad applicability. 
QK-Edit represents a pivotal step in bridging the gap between the generative power and editable flexibility of Transformer-based diffusion models.

\section{Related Work}
\label{sec:related_work}

\noindent \textbf{From U-Net to MM-DiT in Diffusion Models.}
We revisit diffusion models, which underpin most recent editing methods. The U-Net architecture~\cite{ronneberger2015u, podell2023sdxl} has been extensively studied in generative models, as seen in Lumiere~\cite{bar2024lumiere} for text-to-video synthesis and Stable Diffusion~\cite{rombach2022high} for text-to-image generation. Recently, Transformer-based diffusion models have gained attention for capturing long-range dependencies and multimodal interactions. For instance, Peebles and Xie~\cite{peebles2023scalable} introduce the Diffusion Transformer~(DiT), replacing U-Net with a Vision Transformer~(ViT) for improved results, while Chen et al.\cite{chen2024gentron} propose GenTron to overcome CNN-based U-Net limitations. Multimodal extensions, such as MM-DiT~\cite{sd3, flux}, integrate text and vision embeddings to enable sophisticated multi-prompt generation tasks~\cite{cai2024ditctrl}. For video generative models, following Sora~\cite{sora}, several works~\cite{genmo2024mochi, kong2024hunyuanvideo,lin2024open} have open-sourced video diffusion models like Hunyuan Video~\cite{kong2024hunyuanvideo} and CogVideoX~\cite{yang2024cogvideox}.
These advances illustrate an evolution from U-Net to Transformer-based frameworks, offering enhanced scalability, flexibility, and performance. Our training-free technique offers a fast and low-cost solution to unlock the inherent editing potential of these MM-DiT-based new-generation models.

\smallskip
\noindent \textbf{Diffusion-based Image Editing.}  Recent image editing advances fall into training-free and training-based approaches. Training-free methods use pre-trained generative models for direct image manipulation, avoiding costly training~\cite{meng2025instructgie,jia2024designedit,brack2024leditspp,ju2024pnp,hertzprompt,avrahami2024stable,wang2024taming,guo2024focus, song2024doubly,li2024zone,huberman2024edit, liu2024towards,yang2023object}. They mainly include inversion-based and direct manipulation-based techniques. Inversion-based methods, such as Null-Text Inversion~\cite{mokady2023null}, StyleDiffusion~\cite{wang2023stylediffusion} and KV Inversion~\cite{huang2023kv} encode images into a generative model’s latent space, where edits adjust latent representations. 
Other methods, such as pix2pix-zero~\cite{parmar2023zero} and SEGA~\cite{brack2023sega}, modify latent features or noise maps using attention and blending, enabling precise and efficient editing. In contrast, training-based approaches~\cite{kawar2023imagic, jia2023taming, zhang2023magicbrush, zhang2024hive,geng2024instructdiffusion,qin2023unicontrol,chen2024artadapter,ng2023dreamcreature,wang2024instantid,li2023blip,xie2023smartbrush,huang2024smartedit,li2024photomaker} enhance editing capabilities by fine-tuning generative models or training auxiliary networks on task-specific datasets~\cite{zhang2023magicbrush,brooks2023instructpix2pix,hui2024hq}. These approaches generally offer stronger semantic fidelity and improved visual quality at the cost of additional computational overhead. Methods like InstructPix2Pix~\cite{brooks2023instructpix2pix} and MagicBrush~\cite{zhang2023magicbrush} employ instruction-based editing by fine-tuning generative models to directly map textual instructions and input images to edited outputs. 
Our QK-Edit, built on MM-DiT architectures, requires no extra training while efficiently unlocking new-generation models’ editing potential for high prompt fidelity and structural consistency. 




\smallskip
\noindent
\textbf{Diffusion-based Video Editing.} 
Diffusion-based Video Editing. Training-free methods~\cite{jia2024designedit, bai2024uniedit, liu2024video, wang2023zero, qi2023fatezero, shin2024edit} leverage pre-trained diffusion models and latent manipulation to avoid extensive training. For example, Video-P2P~\cite{liu2024video} and Vid2Vid-Zero~\cite{wang2023zero} extend the image-based P2P~\cite{hertzprompt} framework to video by injecting attention maps and using latent inversion for temporal consistency, while Fate-Zero~\cite{qi2023fatezero} enhances spatial coherence by selectively injecting SA and CA maps from the reconstruction branch. Similarly, Edit-A-Video~\cite{shin2024edit} and Make-A-Protagonist~\cite{zhao2023make} refine latent manipulations with explicit semantic masks or advanced attention mechanisms, improving temporal stability without extra training. In contrast, training-based approaches~\cite{feng2024ccedit,singer2024video,zuo2025cut,liu2024stablev2v, blattmann2023stable, chengconsistent} fine-tune models specifically for video data~\cite{chengconsistent,gu2024edit3k, argaw2022anatomy, hu2024vivid, feng2024ccedit}, significantly boosting editing performance at the cost of increased computation. Tune-A-Video~\cite{wu2023tune} fine-tunes spatial and temporal attention layers for consistency, while MotionDirector~\cite{zhao2024motiondirector} and Dreamix~\cite{molad2023dreamix} employ decoupled motion and appearance fine-tuning to reduce artifacts. ControlNet-based methods, such as MagicEdit~\cite{liew2023magicedit} and Ground-A-Video~\cite{ye2024ground}, use depth maps and bounding boxes to guide edits for improved semantic and temporal coherence, and InsV2~\cite{chengconsistent} leverages motion compensation with instruction-based fine-tuning to handle longer videos with enhanced continuity.
Our QK-Edit can easily generalize to video editing for MM-DiT-based video generative models~\cite{kong2024hunyuanvideo, yang2024cogvideox}.

\section{Method}

We aim to design a general method to edit images and videos based on text prompts for MM-DiTs models~\cite{flux, kong2024hunyuanvideo, kuanyv2v}.
Given a source image $I^{src}$ or video $V^{src}$ to edit, a source prompt $P^{src}$ describing the source vision content, and a target prompt $P^{tgt}$ describing the desired editing target, our goal is to generate a modified image $I^{tgt}$ or video $V^{tgt}$ that exhibits the desired changes specified by $P^{tgt}$ while preserving the unedited region.
To achieve it, our QK-Edit leverages the vision part of the query and key embeddings in the Multimodal Diffusion Transformer~(MM-DiT)~\cite{esser2024scaling}. 
For synthesized images and videos, the query and key embeddings are preserved throughout the synthesis process, whereas for real images and videos, these embeddings are derived from the inversion process.


\subsection{Preliminaries}

\noindent \textbf{U-Net-based diffusion models~\cite{rombach2022high, bar2024lumiere}} employ an encoder-decoder framework with skip connections, where text-vision cross-attention and vision-vision self-attention are applied sequentially. 
In particular, self-attention in Eq.~\ref{eq:vision_self_attention} facilitates the pixel-wise dependencies within the image, ensuring the ability to control spatial relationships:
\begin{equation}
\text{Self-Attention} = \text{Softmax}\Big(\frac{Q_{\text{vision}} K_{\text{vision}}^\top}{\sqrt{d}}\Big) V_{\text{vision}}.
\label{eq:vision_self_attention}
\end{equation}
On the other hand, cross-attention (Eq.~\ref{eq:vision_text_attention}) enables the incorporation of external guidance, such as textual prompts:
\begin{equation}
\text{Cross-Attention} = \text{Softmax}\Big(\frac{Q_{\text{vision}} K_{ \text{text}}^\top}{\sqrt{d}}\Big) V_{\text{text}}.
\label{eq:vision_text_attention}
\end{equation}
Existing editing methods typically make use of the modular nature of U-Net architecture for editing. For example, P2P~\cite{hertzprompt} edits images by replacing cross-attention maps, allowing for semantic changes guided by new textual prompts while preserving the original layout. 
PnP~\cite{Tumanyan_2023_CVPR} injects the features and self-attention maps at specific denoising steps with those from a reference image to encourage the generation of images that closely resemble the reference image. 

\smallskip \noindent \textbf{Multimodmal Diffusion Transformers (MM-DiTs)}, on the contrary, as implemented in models such as FLUX~\cite{flux} and HunyuanVideo~\cite{kong2024hunyuanvideo}, adopt a text-vision Multimodal Self-Attention~(MM-SA) mechanism. In this setup, text and visual inputs are concatenated into a unified embedding space, and a single self-attention operation is performed over this combined representation. Consequently, text-vision cross-attention and vision-vision self-attention are integrated within the same attention map, influencing each other through the softmax operation. Mathematically, the MM-SA operation is formulated as
\begin{equation}
\text{MM-SA} = \text{Softmax}\Big(\frac{Q K^\top}{\sqrt{d}}\Big) V,
\end{equation}
where $Q$, $K$ and $V$ are the concatenated query, key and value matrices of text and vision embeddings, respectively. $[\cdot ]$ denotes concatenation operation. 
\begin{equation}
Q = [Q_{\text{text}}, Q_{\text{vision}}], \; K = [K_{\text{text}}, K_{\text{vision}}], \; V = [V_{\text{text}}, V_{\text{vision}}],
\end{equation}


This integrated approach enables more nuanced interactions between textual and visual modalities, enhancing the model capability to generate coherent and contextually relevant outputs. However, it also introduces complexity, as the intertwined attention mechanisms can lead to intricate dependencies that affect the interpretability and controllability of the model behavior.

\subsection{Analysis of MM-DiT}

Taking FLUX~\cite{flux} as an example, if we directly apply the cross-attention and self-attention manipulation techniques, the results are not as satisfactory as in U-Net diffusion~\cite{hertzprompt, Tumanyan_2023_CVPR}, as shown in Fig.\ref{fig:Sec3.1_DirectUseFailure}. Understanding these distinctions is crucial for developing methods that leverage the strengths of each architecture, particularly in tasks requiring precise control over the generated content.

Next, we conduct a progressive analysis of the model, from the main modules to the finer details, to fully understand the design and functionality of MM-DiT. In the following experiments, we take FLUX~\cite{flux} and HunyuanVideo~\cite{kong2024hunyuanvideo} as representative image and video generative models, respectively.

\begin{figure}[t]
    \centering
    \includegraphics[width=\columnwidth]{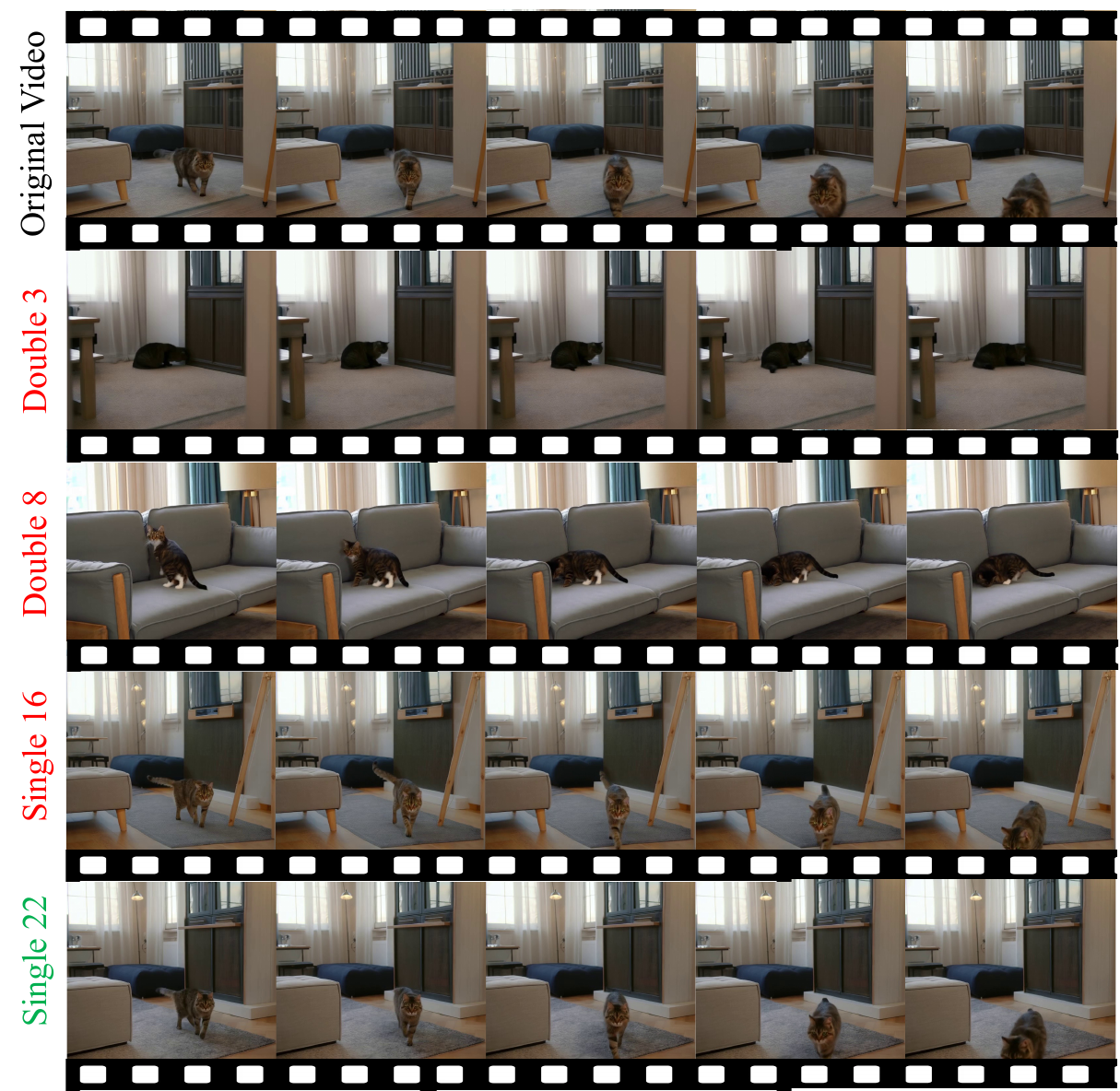}
    \caption{Removing individual blocks in the MM-DiT architecture affects different aspects of the generated video.
    }
    \label{fig:Sec3.2_layers}
\end{figure}

\smallskip \noindent \textbf{Principal and non-principal blocks. }
We first examine the role of each MM-DiT block on HunyuanVideo~\cite{kong2024hunyuanvideo}. Given the same prompt, we skip one block at a time and observed that skipping certain blocks significantly affects the generated videos, while others have different or minimal impacts. For example, as shown in Fig.~\ref{fig:Sec3.2_layers}, skipping the $22^{nd}$ single block does not lead to visible change, whereas bypassing other blocks causes significant changes. This phenomenon is also consistent with the observations in Stable Flow~\cite{avrahami2024stable}. Next, we investigate these blocks to better understand the underlying mechanism.


\begin{figure}[t]
    \centering
    \includegraphics[width=\columnwidth]{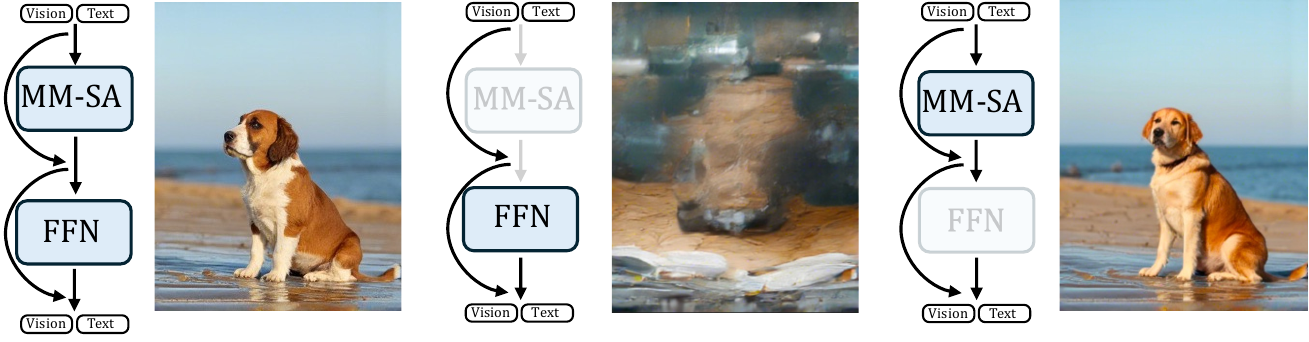}
    \caption{We study the role of the Multimodal Self-Attention (MM-SA) and Feed-Forward Network (FFN) components in MM-DiT blocks by bypassing one at the time. The results underscore that MM-SA predominantly determines spatial and temporal consistency, while FFN has a negligible impact.}
    \label{fig:Sec3.2_attention_ffn}
\end{figure}


\smallskip \noindent \textbf{MM-SA decides the final spatial arrangement. }
Each MM-DiT block comprises a MM-SA and a Feed-forward Network (FFN), connected through residual connections. As before, we skip one component at a time to inspect its impact on the final generation. As shown in Fig.~\ref{fig:Sec3.2_attention_ffn}, MM-SA decides the final spatial arrangement, while skipping FFN has negligible effect. This inspires us to focus on the MM-SA for potential editing. Next, we analyze each attention quadrant of the MM-SA attention map.






\begin{figure*}[t]
    \centering
    \includegraphics[width=\textwidth]{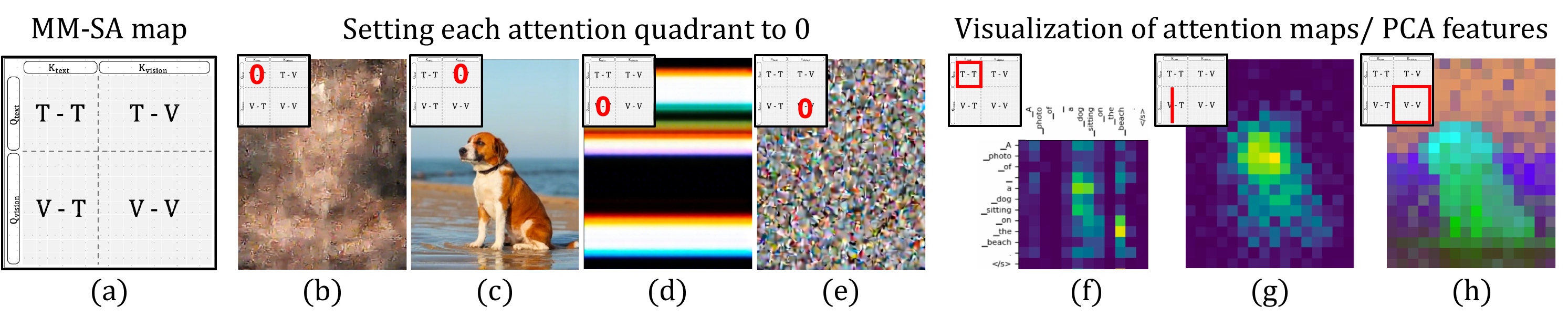}
    \caption{\textbf{Pilot experiments.} (left) MM-SA attention map. (middle) prompting the influence of each attention quadrant by setting it to $\mathbf{0}$. (right) visualization of attention map of the \nth{1} and \nth{3} quadrants and PCA feature of the \nth{4} quadrant.}
    \label{fig:Sec3.3_Pilot}
\end{figure*}



\smallskip \noindent \textbf{Attention quadrants.}
The attention map in MM-SA can be decomposed into 4 quadrants as shown in Fig.~\ref{fig:Sec3.3_Pilot}(a). 
Based on this, we conduct two pilot studies: in Fig.~\ref{fig:Sec3.3_Pilot}(b), (c), (d), and (e), we analyze the effect of each attention quadrant by setting its values to 0; and in Fig.~\ref{fig:Sec3.3_Pilot}(f), (g), and (h), we visualize the attention map of the \nth{1} and \nth{3} quadrants and the PCA feature of the \nth{4} quadrant. 


\smallskip
\noindent \textbf{(i) \nth{1} Quadrant}: $Q_{ \text{text}} K_{ \text{text}}^\top$ controls the relationship among the text tokens of prompt $P^{j}, \forall j \in \{ { src}, { tgt} \}$ as shown in  Fig.~\ref{fig:Sec3.3_Pilot}(f). It is not symmetrical. And the attention value in this quadrant is usually $\gg$ the other 3 vision-related counterparts.

\smallskip
\noindent \textbf{(ii) \nth{2} Quadrant}: $Q_{\text{text}} K_{\text{vision}}^\top$ is not as important as the others, although both the \nth{2} and \nth{3} Quadrant represent the text-vision relationship in theory. When removing this quadrant~(\ie setting this quadrant to 0), there is a negligible effect in the generated result (Fig.~\ref{fig:Sec3.3_Pilot}(c)).

\smallskip
\noindent \textbf{(iii) \nth{3} Quadrant}: $Q_{ \text{vision}} K_{ \text{text}}^\top$ is responsible for deciding the text-vision relationship. Different from the \nth{2} Quadrant, FLUX fails to generate a proper image when setting the 3rd Quadrant to 0 (see Fig.~\ref{fig:Sec3.3_Pilot}(d)). Every column is related to the region corresponding to a certain text token as shown in Fig.~\ref{fig:Sec3.3_Pilot}(g). 

\smallskip
\noindent \textbf{(iv) \nth{4} Quadrant}: $Q_{ \text{vision}} K_{ \text{vision}}^\top$ is responsible for deciding the spatial layout since it defines the pixel affinity (similarity between pixel $i$ and $j$) (Fig.~\ref{fig:Sec3.3_Pilot}(h).

\subsection{Editing using Multimodal Self-Attention}
In the previous section, we empirically found that the MM-SA determines the spatial arrangement~(spatial layout in image, motion in video). Therefore, to achieve desirable editing, we need to ``fix'' the elements in MM-SA to preserve the unchanged elements in the source video $V^{src}$ and modify the region to change following the target prompt $P^{tgt}$.

\begin{figure}[h]
    \centering
    \includegraphics[width=\columnwidth]{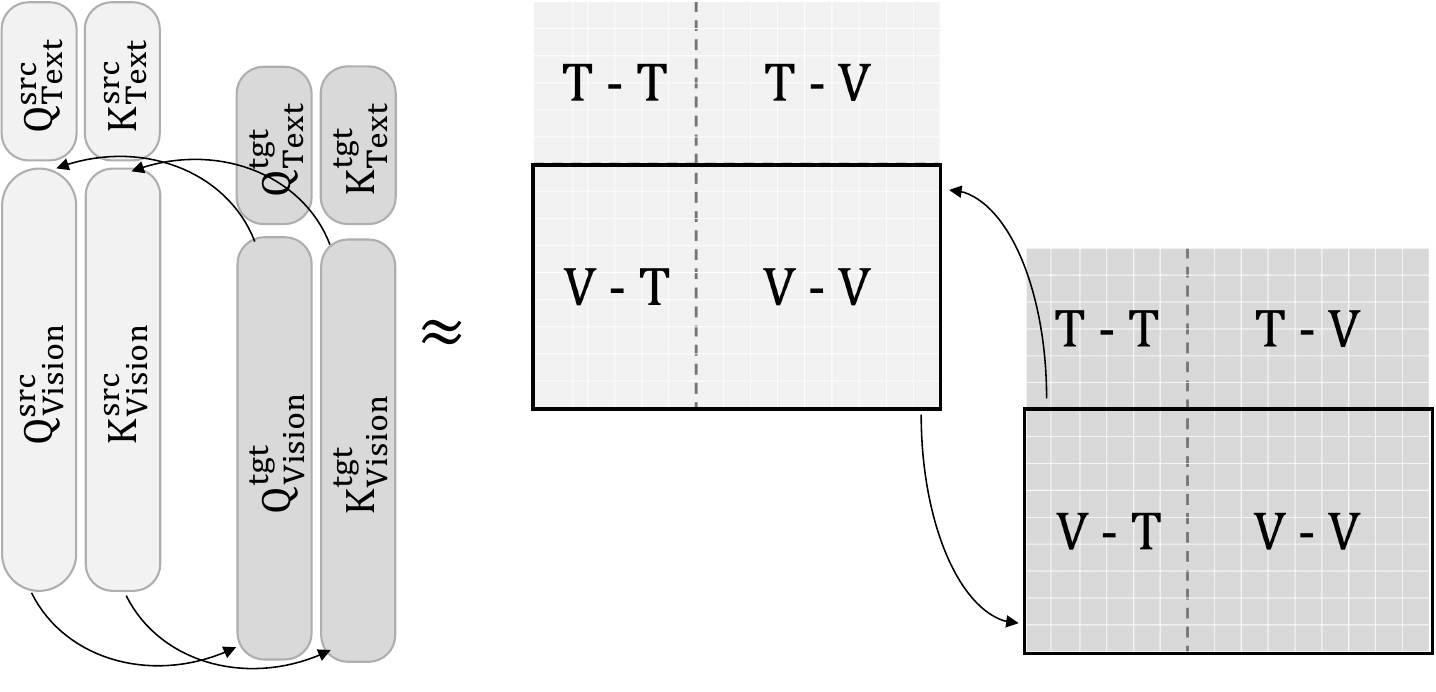}
    \caption{\textbf{Query-key manipulation strategy.} By substituting the vision components in Q and K, the approach effectively guides the editing process to balance prompt fidelity with preservation of the source structure.}
    \label{fig:Sec3.5_replace_2}
\end{figure}

Based on these observations, we show that text-guided editing can be performed by manipulating \textbf{\nth{3} Quadrant} and \textbf{\nth{4} Quadrant}, \ie ${Q_{ \text{vision}} K_{ \text{text}}^\top}$ and $Q_{ \text{vision}} K_{ \text{vision}}^\top$. Specifically, we begin with a source prompt $P^{src}$ and a source video $V^{src}$ to edit. In the process of generating source video from source prompt or inversing real video to initial noise, we record \textcolor{deepgreen}{$Q_{ \text{vision}}^{ src}$} and \textcolor{deepgreen}{$K_{ \text{vision}}^{ src \top}$} of selected time steps. To edit the source video according to the target prompt, 
we manipulate the \nth{3} and \nth{4} Quadrant during the denoising process by:
\begin{equation}
Q_{ \text{vision}}^{ tgt} K_{ \text{text}}^{ tgt \top} \leftarrow \mathcolor{deepgreen}{Q_{ \text{vision}}^{ src}} K_{ \text{text}}^{ tgt \top},
\end{equation}
\begin{equation}
Q_{ \text{vision}}^{ tgt} K_{ \text{vision}}^{ tgt \top} \leftarrow \mathcolor{deepgreen}{Q_{ vision}^{ src} K_{ \text{vision}}^{ src \top}}.
\end{equation}
However, in practice, attention operation is implemented in a highly optimized and integrated manner, where attention maps are often encapsulated within a package (\eg, {\tt flash-attention}\footnote{\url{https://github.com/Dao-AILab/flash-attention}}). As a result, it is challenging to directly modify the attention maps.

Building on our previous observation that \nth{2} Quadrant has little impact on the final output, we can circumvent the constraints imposed by this encapsulation. Consequently, manipulating \nth{3} and \nth{4} Quadrant becomes equivalent to substituting the vision component of \(Q^{tgt}\) and \(K^{tgt}\), i.e.
\begin{equation}
[Q_{ \text{text}}^{ tgt}, Q_{ \text{vision}}^{ tgt}] \leftarrow [Q_{ \text{text}}^{ tgt}, \textcolor{deepgreen}{Q_{ \text{vision}}^{ src}}],
\end{equation}
\begin{equation}
[K_{ \text{text}}^{ tgt}, K_{ \text{vision}}^{ tgt}] \leftarrow [K_{ \text{text}}^{ tgt}, \textcolor{deepgreen}{K_{ \text{vision}}^{ src}}].
\end{equation}
This substitution is illustrated in Fig.~\ref{fig:Sec3.5_replace_2}. Finally, the manipulated QKV is used to generate the target image/ video. 

\smallskip
\noindent \textbf{Easy Implementation.} 
QK-Edit is straightforward to implement. It involves only two steps: (1) during real video inversion or generative video generation, save the $Q_{ \text{vision}}^{ src}$ and $K_{ \text{vision}}^{ src}$; and (2) during target video generation, use the saved ones to replace the $Q_{ \text{vision}}^{ tgt}$ and $K_{ \text{vision}}^{ tgt}$. 



\section{Experiments}

\subsection{Implementation details}
We implement QK-Edit on state-of-the-art MM-DiT-based generative models. Specifically, we use FLUX.1-dev~\cite{flux} and HunyuanVideo~\cite{kong2024hunyuanvideo} as base model for image and video editing, respectively. 
For image editing, we apply our query-key manipulation during the first 30\% diffusion inference steps. 
For video editing, resolutions vary between 320$\times$320 (with 65 frames) and 1280$\times$720 (up to 129 frames). We apply query-key manipulation during the first 20\% of inference steps, which empirically provides optimal temporal consistency and editing fidelity. All experiments are conducted on NVIDIA A100 GPUs.

\subsection{Quantitative Evaluation Metrics}
To rigorously assess editing performance, we utilize the following metrics following previous works~\cite{xu2024headrouter, kirstain2023pick, singer2024video, feng2024ccedit}. 


Specifically, for image editing, we employ \textbf{CLIP$_{\text{text}}$} (measures semantic alignment between the edited image and the text prompt), \textbf{DINO} (perceptual similarity for structure-alignment) and \textbf{PickScore} (assesses overall editing quality and realism in a human-like manner).

On the other hand, to evaluate video editing performance, we use \textbf{CLIP$_{\text{frames}}$} (computes frame consistency through the average cosine similarity across all video frames), \textbf{PickScore} (assesses the predicted human preference averaged over all video frames), \textbf{ViCLIP$_{\text{dir}}$}(measures directional video-text alignment) and \textbf{ViCLIP$_{\text{out}}$} (evaluates overall video-text alignment).


\subsection{Quantitative Results}

\begin{table}[t]
\centering
{\small
\renewcommand{\arraystretch}{0.8}
\begin{tabular}{l|ccc}
    \toprule
    Method & CLIP-T↑ & DINO↑ & PickScore↑ \\
    \midrule
    LEDITS++~\cite{brack2024leditspp}  & 0.207  & 0.571  & 19.88 \\
    MasaCtrl~\cite{cao2023masactrl}    & 0.235  & 0.696  & 20.68 \\
    RF-Inversion~\cite{rout2024semantic} & 0.252  & 0.523  & 21.31 \\
    RF-Solver~\cite{wang2024taming}    & 0.257  & 0.769  & 21.98 \\
\rowcolor{lightgray} QK-Edit       & \textbf{0.264} & \textbf{0.776} & \textbf{22.33} \\
    \bottomrule
\end{tabular}
}
\caption{\textbf{Real image editing evaluation on PIE-Bench}. 
}
\label{tab:image_quantity}
\end{table}


\noindent \textbf{Real image editing.} For image editing, we compare our QK-Edit against LEDITS++~\cite{brack2024leditspp}, MasaCtrl~\cite{cao2023masactrl}, RF-Inversion~\cite{rout2024semantic}, and RF-Solver~\cite{wang2024taming} on the PIE-Bench~\cite{jupnp} (Tab.~\ref{tab:image_quantity}). As shown in Tab.~\ref{tab:image_quantity}, QK-Edit achieves the highest scores in all metrics, indicating its superiority in aligning with the text prompt while preserving structural details. 

\begin{figure*}[t]
    \centering
    \includegraphics[width=0.99\linewidth]{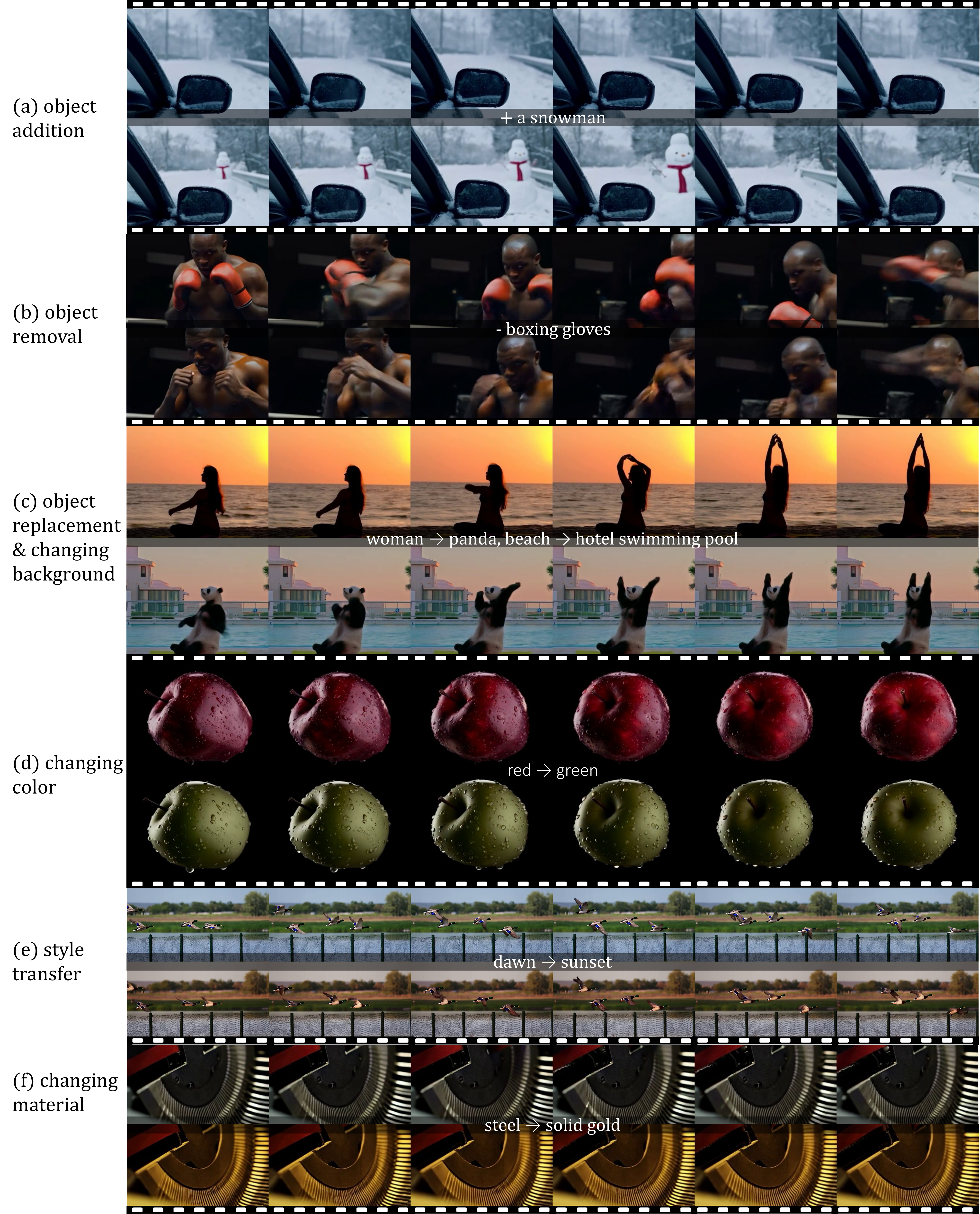}
    \vspace{-2mm}
    \caption{Visualizations showcasing QK-Edit’s capabilities in various video editing tasks.}
    \label{fig:Sec4.4_vis}
\end{figure*}

\smallskip
\noindent \textbf{Video editing.} For video editing, we evaluate on the TGVE+ benchmark~\cite{singer2024video} with two settings. In the first setting, we evaluate on the generative video editing task (without video inversion. Instead, source video is generated as the first step). We compare our QK-Edit with DiTCtrl~\cite{cai2024ditctrl}. Both methods are based on HunyuanVideo~\cite{kong2024hunyuanvideo} for fair comparison. As shown in Tab.~\ref{tab:video_syn_quantity}, QK-Edit outperforms DiTCtrl across all metrics, demonstrating better frame consistency and stronger video-text alignment.


In the second setting, we evaluate the performance of real video editing (requires video inversion as the first step). We evaluate QK-Edit against CCEdit~\cite{feng2024ccedit}, Flatten~\cite{congflatten}, and AnyV2V~\cite{kuanyv2v}. As shown in Tab.~\ref{tab:video_real_quantity}, QK-Edit outperforms all the baselines, validating that QK-Edit not only produces visually coherent edits but also maintains semantic consistency across video frames.



\begin{table}[t]
\centering
\footnotesize
\renewcommand\arraystretch{1.0}
\resizebox{\columnwidth}{!}{%
\begin{tabular}{l|cccc}
    \toprule
    Method & CLIP$_{\texttt{frames}}$↑ & PickScore↑ & ViCLIP$_{\texttt{dir}}$↑ & ViCLIP$_{\texttt{out}}$↑ \\
    \midrule
    DiTCtrl~\cite{cai2024ditctrl} & 0.904 & 20.01 & 0.141 & 0.242 \\
\rowcolor{lightgray} QK-Edit & \textbf{0.951} & \textbf{21.24} & \textbf{0.217} & \textbf{0.268} \\
    \bottomrule
\end{tabular}%
}
\caption{\textbf{Generative video editing performance on TGVE+ dataset.} 
}
\label{tab:video_syn_quantity}
\end{table}

\begin{table}[t]
\centering
\footnotesize
\renewcommand\arraystretch{1.0}
\resizebox{\columnwidth}{!}{%
\begin{tabular}{l|cccc}
    \toprule
    Method & ViCLIP$_\texttt{dir}$↑ & ViCLIP$_\texttt{out}$↑ & CLIP$_\texttt{frames}$↑ & PickScore↑ \\
    \midrule
    CCEdit~\cite{feng2024ccedit}   & 0.159 & 0.244 & 0.877 & 19.73 \\
    Flatten~\cite{congflatten}       & 0.204 & 0.236 & 0.887 & 19.91 \\
    AnyV2V~\cite{kuanyv2v}           & 0.117 & 0.228 & 0.906 & 20.28 \\
\rowcolor{lightgray} QK-Edit      & \textbf{0.196} & \textbf{0.258} & \textbf{0.921} & \textbf{20.52} \\
    \bottomrule
\end{tabular}%
}
\caption{\textbf{Real video editing performance on TGVE+ dataset.} 
}
\label{tab:video_real_quantity}
\end{table}

\subsection{Qualitative Comparisons}
In addition to the quantitative evaluation, we also present some qualitative comparisons. As shown in Fig.~\ref{fig:Sec4.4_vis}, we can see that QK-Edit not only preserves fine-grained details and structures, but also is capable to integrate target textual cues into the edited image.
On the other hand, for video editing, QK-Edit maintains excellent frame-to-frame consistency and temporal continuity. These visual examples underscore the method’s ability to deliver high-fidelity edits while remaining temporally coherent.

Besides, we also found that QK-Edit excels in multiple editing scenario, including
object addition (\eg, add a snowman in Fig.~\ref{fig:Sec4.4_vis}(a)), object removal (\eg, remove the boxing gloves in Fig.~\ref{fig:Sec4.4_vis}(b)), object replacement (\eg,  woman to panda in Fig.~\ref{fig:Sec4.4_vis}(c)), changing background (\eg, beach to hotel swimming pool in Fig.~\ref{fig:Sec4.4_vis}(c)), changing color (\eg, red to green in Fig.~\ref{fig:Sec4.4_vis}(d)), style transformation (\eg, from dawn to sunset in Fig.~\ref{fig:Sec4.4_vis}(e)), and changing material (\eg, from steel to solid gold in Fig.~\ref{fig:Sec4.4_vis}(f)). These examples demonstrate the flexibility of QK-Edit across a wide range of editing tasks.




\subsection{Ablation Study}

We perform extensive ablation studies to validate our design choices and more challenging scenario.

\begin{figure}[t]
    \centering
    \includegraphics[width=\columnwidth]{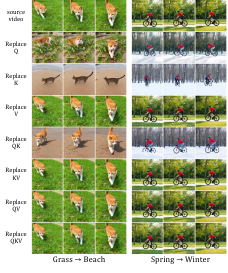}
    \caption{\textbf{Manipulation of different QKV combinations.} Replacing $KV$ or $V$ preserves original structure better, but sometimes fails to follow the target prompt; replacing $Q$ or $K$, on the hand, adheres to the editing prompt but often fail to preserve source video structure; replacing $QK$ achieves the best trade-off.}
    \label{fig:ablation_QKV}
\end{figure}

\smallskip
\noindent {\bf Manipulation on QKV combinations.}
We perform an ablation study to understand the effect of utilizing different combinations of $Q_{\text{vision}}$, $K_{\text{vision}}$ and $V_{\text{vision}}$ in Fig~\ref{fig:ablation_QKV}. We observe that (1) replacing $QK$ works the best in preserving the source video structure while following the target prompt, (2) replacing $KV$ or $V$ preserve source video structure better, but sometimes fails to follow the target prompt. 
(3) replacing $Q$ or $K$ can better follow the prompt, but often fails to preserve the source video structure.



\smallskip \noindent \textbf{High-fidelity video editing.} 
Recognizing the growing demand for high-fidelity video editing, we introduce a new benchmark, HiLo-Benchmark, consisting of 43 carefully selected challenging videos with resolutions exceeding 1K and sequences longer than 100 frames. This benchmark specifically evaluates the ability of video editing methods to handle extended and high-resolution content.
For baselines, considering compatibility with high-resolution and long-duration video generation, we adapt ideas from previous image/video synthesis frameworks and reimplement them based on HunyuanVideo~\cite{kong2024hunyuanvideo}. Our proposed benchmark includes comparisons with methods such as KV$_{\text{concat}}$ (MasaCtrl~\cite{cao2023masactrl} and Tune-a-video~\cite{wu2023tune}), V$_{\text{replace}}$ (RF-Solver~\cite{wang2024taming}) and KV$_{\text{replace}}$ (Stable Flow~\cite{avrahami2024stable}).
Tab.~\ref{tab:HiLo} demonstrates that QK-Edit consistently achieves higher editing fidelity, structural consistency, and semantic alignment across both images and videos compared to existing approaches.


\begin{table}[t]
\centering
\footnotesize
\renewcommand\arraystretch{1.0}
{
\begin{tabular}{l|ccc>{\columncolor{lightgray}}c}
    \toprule
    Method & KV$_{\text{concat}}$ & V$_{\text{replace}}$ & KV$_{\text{replace}}$ & QK$_{\text{replace}}$ \\
    \midrule
    CLIP$_\texttt{frames}$↑ & 0.896 & 0.900 & 0.901 &  \textbf{0.908} \\
    PickScore↑ & 19.87 &  19.94 & 19.92 &  \textbf{20.08} \\
\bottomrule
\end{tabular}
}
\caption{\textbf{Quantitative results on HiLo-Benchmark}. Our proposed QK-Edit via QK$_{\text{replace}}$ method (last column) works the best.}
\label{tab:HiLo}
\end{table}

\section{Conclusion}

In this work, we present QK-Edit, a simple yet effective training-free framework that exploits the unique attention dynamics of MM-DiTs for precise text-guided image and video editing. Extensive experiments demonstrate the versatility of QK-Edit across a wide range of editing tasks, \eg, object addition, object removal, object replacement, changing background, changing material, changing color, and style transformation \etc. More importantly, it can be easily implemented with a few lines of code.

\smallskip \noindent
\textbf{Limitation and future work.}
While QK-Edit significantly advances training-free text-guided editing for MM-DiTs, it still exhibits several limitations. First, the accuracy of real video inversion remains challenging, particularly in complex scenes with intricate motions, which can compromise subsequent editing results. Second, QK-Edit inevitably introduces minor unintended modifications in regions that should remain unaltered; future work could incorporate masks to better preserve these areas. Third, as a training-free approach, QK-Edit inherits the limitations of its base generative model, such as suboptimal text rendering and structural deficiencies in handling complex motions (e.g., gymnastic videos). These issues might be alleviated by employing more advanced generative models, which we leave as a promising direction for future research.




{
    \small
    \bibliographystyle{ieeenat_fullname}
    \bibliography{main}

\begin{thebibliography}{76}
\providecommand{\natexlab}[1]{#1}
\providecommand{\url}[1]{\texttt{#1}}
\expandafter\ifx\csname urlstyle\endcsname\relax
  \providecommand{\doi}[1]{doi: #1}\else
  \providecommand{\doi}{doi: \begingroup \urlstyle{rm}\Url}\fi

\bibitem[Argaw et~al.(2022)Argaw, Heilbron, Lee, Woodson, and
  Kweon]{argaw2022anatomy}
Dawit~Mureja Argaw, Fabian~Caba Heilbron, Joon-Young Lee, Markus Woodson, and
  In~So Kweon.
\newblock The anatomy of video editing: A dataset and benchmark suite for
  ai-assisted video editing.
\newblock In \emph{ECCV}, 2022.

\bibitem[Avrahami et~al.(2024)Avrahami, Patashnik, Fried, Nemchinov, Aberman,
  Lischinski, and Cohen-Or]{avrahami2024stable}
Omri Avrahami, Or Patashnik, Ohad Fried, Egor Nemchinov, Kfir Aberman, Dani
  Lischinski, and Daniel Cohen-Or.
\newblock Stable flow: Vital layers for training-free image editing.
\newblock \emph{arXiv preprint arXiv:2411.14430}, 2024.

\bibitem[Bai et~al.(2024)Bai, He, Wang, Guo, Hu, Liu, and Bian]{bai2024uniedit}
Jianhong Bai, Tianyu He, Yuchi Wang, Junliang Guo, Haoji Hu, Zuozhu Liu, and
  Jiang Bian.
\newblock Uniedit: A unified tuning-free framework for video motion and
  appearance editing.
\newblock \emph{arXiv preprint arXiv:2402.13185}, 2024.

\bibitem[Bar-Tal et~al.(2024)Bar-Tal, Chefer, Tov, Herrmann, Paiss, Zada,
  Ephrat, Hur, Liu, Raj, et~al.]{bar2024lumiere}
Omer Bar-Tal, Hila Chefer, Omer Tov, Charles Herrmann, Roni Paiss, Shiran Zada,
  Ariel Ephrat, Junhwa Hur, Guanghui Liu, Amit Raj, et~al.
\newblock Lumiere: A space-time diffusion model for video generation.
\newblock In \emph{SIGGRAPH Asia}, 2024.

\bibitem[{Black Forest Labs}(2023)]{flux}
{Black Forest Labs}.
\newblock The announcements page of black forest labs website.
\newblock \url{https://blackforestlabs.ai/announcing-black-forest-labs/}, 2023.
\newblock Accessed: 2024-4.

\bibitem[Blattmann et~al.(2023)Blattmann, Dockhorn, Kulal, Mendelevitch,
  Kilian, Lorenz, Levi, English, Voleti, Letts, et~al.]{blattmann2023stable}
Andreas Blattmann, Tim Dockhorn, Sumith Kulal, Daniel Mendelevitch, Maciej
  Kilian, Dominik Lorenz, Yam Levi, Zion English, Vikram Voleti, Adam Letts,
  et~al.
\newblock Stable video diffusion: Scaling latent video diffusion models to
  large datasets.
\newblock \emph{arXiv preprint arXiv:2311.15127}, 2023.

\bibitem[Brack et~al.(2023)Brack, Friedrich, Hintersdorf, Struppek,
  Schramowski, and Kersting]{brack2023sega}
Manuel Brack, Felix Friedrich, Dominik Hintersdorf, Lukas Struppek, Patrick
  Schramowski, and Kristian Kersting.
\newblock Sega: Instructing text-to-image models using semantic guidance.
\newblock \emph{NeurIPS}, 2023.

\bibitem[Brack et~al.(2024)Brack, Friedrich, Kornmeier, Tsaban, Schramowski,
  Kersting, and Passos]{brack2024leditspp}
Manuel Brack, Felix Friedrich, Katharia Kornmeier, Linoy Tsaban, Patrick
  Schramowski, Kristian Kersting, and Apolin{\'a}rio Passos.
\newblock Ledits++: Limitless image editing using text-to-image models.
\newblock In \emph{CVPR}, 2024.

\bibitem[Brooks et~al.(2023)Brooks, Holynski, and
  Efros]{brooks2023instructpix2pix}
Tim Brooks, Aleksander Holynski, and Alexei~A Efros.
\newblock Instructpix2pix: Learning to follow image editing instructions.
\newblock In \emph{CVPR}, 2023.

\bibitem[Cai et~al.(2024)Cai, Cun, Li, Liu, Zhang, Zhang, Shan, and
  Yue]{cai2024ditctrl}
Minghong Cai, Xiaodong Cun, Xiaoyu Li, Wenze Liu, Zhaoyang Zhang, Yong Zhang,
  Ying Shan, and Xiangyu Yue.
\newblock Ditctrl: Exploring attention control in multi-modal diffusion
  transformer for tuning-free multi-prompt longer video generation.
\newblock \emph{arXiv preprint arXiv:2412.18597}, 2024.

\bibitem[Cao et~al.(2023)Cao, Wang, Qi, Shan, Qie, and Zheng]{cao2023masactrl}
Mingdeng Cao, Xintao Wang, Zhongang Qi, Ying Shan, Xiaohu Qie, and Yinqiang
  Zheng.
\newblock Masactrl: Tuning-free mutual self-attention control for consistent
  image synthesis and editing.
\newblock In \emph{ICCV}, 2023.

\bibitem[Chen et~al.(2024{\natexlab{a}})Chen, Tennent, and
  Hsu]{chen2024artadapter}
Dar-Yen Chen, Hamish Tennent, and Ching-Wen Hsu.
\newblock Artadapter: Text-to-image style transfer using multi-level style
  encoder and explicit adaptation.
\newblock In \emph{CVPR}, 2024{\natexlab{a}}.

\bibitem[Chen et~al.(2024{\natexlab{b}})Chen, Xu, Ren, Cong, He, Xie, Sinha,
  Luo, Xiang, and Perez-Rua]{chen2024gentron}
Shoufa Chen, Mengmeng Xu, Jiawei Ren, Yuren Cong, Sen He, Yanping Xie, Animesh
  Sinha, Ping Luo, Tao Xiang, and Juan-Manuel Perez-Rua.
\newblock Gentron: Diffusion transformers for image and video generation.
\newblock In \emph{CVPR}, 2024{\natexlab{b}}.

\bibitem[Cheng et~al.(2024)Cheng, Xiao, and He]{chengconsistent}
Jiaxin Cheng, Tianjun Xiao, and Tong He.
\newblock Consistent video-to-video transfer using synthetic dataset.
\newblock In \emph{ICLR}, 2024.

\bibitem[Cong et~al.(2024)Cong, Xu, Chen, Ren, Xie, Perez-Rua, Rosenhahn,
  Xiang, He, et~al.]{congflatten}
Yuren Cong, Mengmeng Xu, Shoufa Chen, Jiawei Ren, Yanping Xie, Juan-Manuel
  Perez-Rua, Bodo Rosenhahn, Tao Xiang, Sen He, et~al.
\newblock Flatten: optical flow-guided attention for consistent text-to-video
  editing.
\newblock In \emph{The Twelfth International Conference on Learning
  Representations}, 2024.

\bibitem[Esser et~al.(2024{\natexlab{a}})Esser, Kulal, Blattmann, Entezari,
  M\"{u}ller, Saini, Levi, Lorenz, Sauer, Boesel, Podell, Dockhorn, English,
  and Rombach]{sd3}
Patrick Esser, Sumith Kulal, Andreas Blattmann, Rahim Entezari, Jonas
  M\"{u}ller, Harry Saini, Yam Levi, Dominik Lorenz, Axel Sauer, Frederic
  Boesel, Dustin Podell, Tim Dockhorn, Zion English, and Robin Rombach.
\newblock Scaling rectified flow transformers for high-resolution image
  synthesis.
\newblock In \emph{ICML}, 2024{\natexlab{a}}.

\bibitem[Esser et~al.(2024{\natexlab{b}})Esser, Kulal, Blattmann, Entezari,
  M{\"u}ller, Saini, Levi, Lorenz, Sauer, Boesel, et~al.]{esser2024scaling}
Patrick Esser, Sumith Kulal, Andreas Blattmann, Rahim Entezari, Jonas
  M{\"u}ller, Harry Saini, Yam Levi, Dominik Lorenz, Axel Sauer, Frederic
  Boesel, et~al.
\newblock Scaling rectified flow transformers for high-resolution image
  synthesis.
\newblock In \emph{ICML}, 2024{\natexlab{b}}.

\bibitem[Feng et~al.(2024)Feng, Weng, Wang, Yuan, Bao, Luo, Chen, and
  Guo]{feng2024ccedit}
Ruoyu Feng, Wenming Weng, Yanhui Wang, Yuhui Yuan, Jianmin Bao, Chong Luo,
  Zhibo Chen, and Baining Guo.
\newblock Ccedit: Creative and controllable video editing via diffusion models.
\newblock In \emph{CVPR}, 2024.

\bibitem[Geng et~al.(2024)Geng, Yang, Hang, Li, Gu, Zhang, Bao, Zhang, Li, Hu,
  et~al.]{geng2024instructdiffusion}
Zigang Geng, Binxin Yang, Tiankai Hang, Chen Li, Shuyang Gu, Ting Zhang,
  Jianmin Bao, Zheng Zhang, Houqiang Li, Han Hu, et~al.
\newblock Instructdiffusion: A generalist modeling interface for vision tasks.
\newblock In \emph{CVPR}, 2024.

\bibitem[Gu et~al.(2024)Gu, Zhang, Chen, Wen, Wang, Luo, and Zhu]{gu2024edit3k}
Xin Gu, Libo Zhang, Fan Chen, Longyin Wen, Yufei Wang, Tiejian Luo, and Sijie
  Zhu.
\newblock Edit3k: Universal representation learning for video editing
  components.
\newblock \emph{arXiv preprint arXiv:2403.16048}, 2024.

\bibitem[Guo and Lin(2024)]{guo2024focus}
Qin Guo and Tianwei Lin.
\newblock Focus on your instruction: Fine-grained and multi-instruction image
  editing by attention modulation.
\newblock In \emph{CVPR}, 2024.

\bibitem[Hertz et~al.(2023)Hertz, Mokady, Tenenbaum, Aberman, Pritch, and
  Cohen-or]{hertzprompt}
Amir Hertz, Ron Mokady, Jay Tenenbaum, Kfir Aberman, Yael Pritch, and Daniel
  Cohen-or.
\newblock Prompt-to-prompt image editing with cross-attention control.
\newblock In \emph{ICLR}, 2023.

\bibitem[Hu et~al.(2024)Hu, Zhong, Wang, Jiang, Tian, Yang, Wan, and
  Zhang]{hu2024vivid}
Jiahao Hu, Tianxiong Zhong, Xuebo Wang, Boyuan Jiang, Xingye Tian, Fei Yang,
  Pengfei Wan, and Di Zhang.
\newblock Vivid-10m: A dataset and baseline for versatile and interactive video
  local editing.
\newblock \emph{arXiv preprint arXiv:2411.15260}, 2024.

\bibitem[Huang et~al.(2023)Huang, Liu, Qin, and Chen]{huang2023kv}
Jiancheng Huang, Yifan Liu, Jin Qin, and Shifeng Chen.
\newblock Kv inversion: Kv embeddings learning for text-conditioned real image
  action editing.
\newblock In \emph{PRCV}, 2023.

\bibitem[Huang et~al.(2024)Huang, Xie, Wang, Yuan, Cun, Ge, Zhou, Dong, Huang,
  Zhang, et~al.]{huang2024smartedit}
Yuzhou Huang, Liangbin Xie, Xintao Wang, Ziyang Yuan, Xiaodong Cun, Yixiao Ge,
  Jiantao Zhou, Chao Dong, Rui Huang, Ruimao Zhang, et~al.
\newblock Smartedit: Exploring complex instruction-based image editing with
  multimodal large language models.
\newblock In \emph{CVPR}, 2024.

\bibitem[Huberman-Spiegelglas et~al.(2024)Huberman-Spiegelglas, Kulikov, and
  Michaeli]{huberman2024edit}
Inbar Huberman-Spiegelglas, Vladimir Kulikov, and Tomer Michaeli.
\newblock An edit friendly ddpm noise space: Inversion and manipulations.
\newblock In \emph{CVPR}, 2024.

\bibitem[Hui et~al.(2024)Hui, Yang, Zhao, Shi, Wang, Wang, Zhou, and
  Xie]{hui2024hq}
Mude Hui, Siwei Yang, Bingchen Zhao, Yichun Shi, Heng Wang, Peng Wang, Yuyin
  Zhou, and Cihang Xie.
\newblock Hq-edit: A high-quality dataset for instruction-based image editing.
\newblock \emph{arXiv preprint arXiv:2404.09990}, 2024.

\bibitem[Jia et~al.(2024{\natexlab{a}})Jia, Zhao, Chan, Li, Zhang, Gong, Hou,
  Wang, and Su]{jia2023taming}
Xuhui Jia, Yang Zhao, Kelvin~CK Chan, Yandong Li, Han Zhang, Boqing Gong,
  Tingbo Hou, Huisheng Wang, and Yu-Chuan Su.
\newblock Taming encoder for zero fine-tuning image customization with
  text-to-image diffusion models.
\newblock \emph{ICLR}, 2024{\natexlab{a}}.

\bibitem[Jia et~al.(2024{\natexlab{b}})Jia, Yuan, Cheng, Wang, Li, Jia, and
  Zhang]{jia2024designedit}
Yueru Jia, Yuhui Yuan, Aosong Cheng, Chuke Wang, Ji Li, Huizhu Jia, and
  Shanghang Zhang.
\newblock Designedit: Multi-layered latent decomposition and fusion for unified
  \& accurate image editing.
\newblock \emph{arXiv preprint arXiv:2403.14487}, 2024{\natexlab{b}}.

\bibitem[Ju et~al.(2024{\natexlab{a}})Ju, Zeng, Bian, Liu, and Xu]{ju2024pnp}
Xuan Ju, Ailing Zeng, Yuxuan Bian, Shaoteng Liu, and Qiang Xu.
\newblock Pnp inversion: Boosting diffusion-based editing with 3 lines of code.
\newblock In \emph{ICLR}, 2024{\natexlab{a}}.

\bibitem[Ju et~al.(2024{\natexlab{b}})Ju, Zeng, Bian, Liu, and Xu]{jupnp}
Xuan Ju, Ailing Zeng, Yuxuan Bian, Shaoteng Liu, and Qiang Xu.
\newblock Pnp inversion: Boosting diffusion-based editing with 3 lines of code.
\newblock In \emph{ICLR}, 2024{\natexlab{b}}.

\bibitem[Kawar et~al.(2023)Kawar, Zada, Lang, Tov, Chang, Dekel, Mosseri, and
  Irani]{kawar2023imagic}
Bahjat Kawar, Shiran Zada, Oran Lang, Omer Tov, Huiwen Chang, Tali Dekel, Inbar
  Mosseri, and Michal Irani.
\newblock Imagic: Text-based real image editing with diffusion models.
\newblock In \emph{CVPR}, 2023.

\bibitem[Kirstain et~al.(2023)Kirstain, Polyak, Singer, Matiana, Penna, and
  Levy]{kirstain2023pick}
Yuval Kirstain, Adam Polyak, Uriel Singer, Shahbuland Matiana, Joe Penna, and
  Omer Levy.
\newblock Pick-a-pic: An open dataset of user preferences for text-to-image
  generation.
\newblock \emph{NeurIPS}, 2023.

\bibitem[Kong et~al.(2024)Kong, Tian, Zhang, Min, Dai, Zhou, Xiong, Li, Wu,
  Zhang, et~al.]{kong2024hunyuanvideo}
Weijie Kong, Qi Tian, Zijian Zhang, Rox Min, Zuozhuo Dai, Jin Zhou, Jiangfeng
  Xiong, Xin Li, Bo Wu, Jianwei Zhang, et~al.
\newblock Hunyuanvideo: A systematic framework for large video generative
  models.
\newblock \emph{arXiv preprint arXiv:2412.03603}, 2024.

\bibitem[Ku et~al.(2024)Ku, Wei, Ren, Yang, and Chen]{kuanyv2v}
Max Ku, Cong Wei, Weiming Ren, Huan Yang, and Wenhu Chen.
\newblock Anyv2v: A tuning-free framework for any video-to-video editing tasks.
\newblock \emph{Transactions on Machine Learning Research}, 2024.

\bibitem[Li et~al.(2023)Li, Li, and Hoi]{li2023blip}
Dongxu Li, Junnan Li, and Steven Hoi.
\newblock Blip-diffusion: Pre-trained subject representation for controllable
  text-to-image generation and editing.
\newblock \emph{NeurIPS}, 2023.

\bibitem[Li et~al.(2024{\natexlab{a}})Li, Zeng, Feng, Gao, Liu, Liu, Li, Tang,
  Hu, Liu, et~al.]{li2024zone}
Shanglin Li, Bohan Zeng, Yutang Feng, Sicheng Gao, Xiuhui Liu, Jiaming Liu, Lin
  Li, Xu Tang, Yao Hu, Jianzhuang Liu, et~al.
\newblock Zone: Zero-shot instruction-guided local editing.
\newblock In \emph{CVPR}, 2024{\natexlab{a}}.

\bibitem[Li et~al.(2024{\natexlab{b}})Li, Cao, Wang, Qi, Cheng, and
  Shan]{li2024photomaker}
Zhen Li, Mingdeng Cao, Xintao Wang, Zhongang Qi, Ming-Ming Cheng, and Ying
  Shan.
\newblock Photomaker: Customizing realistic human photos via stacked id
  embedding.
\newblock In \emph{CVPR}, 2024{\natexlab{b}}.

\bibitem[Liew et~al.(2023)Liew, Yan, Zhang, Xu, and Feng]{liew2023magicedit}
Jun~Hao Liew, Hanshu Yan, Jianfeng Zhang, Zhongcong Xu, and Jiashi Feng.
\newblock Magicedit: High-fidelity and temporally coherent video editing.
\newblock \emph{arXiv preprint arXiv:2308.14749}, 2023.

\bibitem[Lin et~al.(2024)Lin, Ge, Cheng, Li, Zhu, Wang, He, Ye, Yuan, Chen,
  et~al.]{lin2024open}
Bin Lin, Yunyang Ge, Xinhua Cheng, Zongjian Li, Bin Zhu, Shaodong Wang, Xianyi
  He, Yang Ye, Shenghai Yuan, Liuhan Chen, et~al.
\newblock Open-sora plan: Open-source large video generation model.
\newblock \emph{arXiv preprint arXiv:2412.00131}, 2024.

\bibitem[Liu et~al.(2024{\natexlab{a}})Liu, Wang, Cao, Jia, and
  Huang]{liu2024towards}
Bingyan Liu, Chengyu Wang, Tingfeng Cao, Kui Jia, and Jun Huang.
\newblock Towards understanding cross and self-attention in stable diffusion
  for text-guided image editing.
\newblock In \emph{CVPR}, 2024{\natexlab{a}}.

\bibitem[Liu et~al.(2024{\natexlab{b}})Liu, Li, Zhang, Lan, and
  Liu]{liu2024stablev2v}
Chang Liu, Rui Li, Kaidong Zhang, Yunwei Lan, and Dong Liu.
\newblock Stablev2v: Stablizing shape consistency in video-to-video editing.
\newblock \emph{arXiv preprint arXiv:2411.11045}, 2024{\natexlab{b}}.

\bibitem[Liu et~al.(2024{\natexlab{c}})Liu, Zhang, Li, Lin, and
  Jia]{liu2024video}
Shaoteng Liu, Yuechen Zhang, Wenbo Li, Zhe Lin, and Jiaya Jia.
\newblock Video-p2p: Video editing with cross-attention control.
\newblock In \emph{CVPR}, 2024{\natexlab{c}}.

\bibitem[Meng et~al.(2025)Meng, Yang, Liu, Tang, Zhao, and
  Wang]{meng2025instructgie}
Zichong Meng, Changdi Yang, Jun Liu, Hao Tang, Pu Zhao, and Yanzhi Wang.
\newblock Instructgie: Towards generalizable image editing.
\newblock In \emph{ECCV}, 2025.

\bibitem[Mokady et~al.(2023)Mokady, Hertz, Aberman, Pritch, and
  Cohen-Or]{mokady2023null}
Ron Mokady, Amir Hertz, Kfir Aberman, Yael Pritch, and Daniel Cohen-Or.
\newblock Null-text inversion for editing real images using guided diffusion
  models.
\newblock In \emph{CVPR}, 2023.

\bibitem[Molad et~al.(2023)Molad, Horwitz, Valevski, Acha, Matias, Pritch,
  Leviathan, and Hoshen]{molad2023dreamix}
Eyal Molad, Eliahu Horwitz, Dani Valevski, Alex~Rav Acha, Yossi Matias, Yael
  Pritch, Yaniv Leviathan, and Yedid Hoshen.
\newblock Dreamix: Video diffusion models are general video editors.
\newblock \emph{arXiv preprint arXiv:2302.01329}, 2023.

\bibitem[Ng et~al.(2023)Ng, Zhu, Song, and Xiang]{ng2023dreamcreature}
Kam~Woh Ng, Xiatian Zhu, Yi-Zhe Song, and Tao Xiang.
\newblock Dreamcreature: Crafting photorealistic virtual creatures from
  imagination.
\newblock \emph{arXiv preprint arXiv:2311.15477}, 2023.

\bibitem[Open-AI(2024)]{sora}
Open-AI.
\newblock Video generation models as world simulators.
\newblock 2024.

\bibitem[Parmar et~al.(2023)Parmar, Kumar~Singh, Zhang, Li, Lu, and
  Zhu]{parmar2023zero}
Gaurav Parmar, Krishna Kumar~Singh, Richard Zhang, Yijun Li, Jingwan Lu, and
  Jun-Yan Zhu.
\newblock Zero-shot image-to-image translation.
\newblock In \emph{SIGGRAPH 2023}, 2023.

\bibitem[Peebles and Xie(2023)]{peebles2023scalable}
William Peebles and Saining Xie.
\newblock Scalable diffusion models with transformers.
\newblock In \emph{ICCV}, 2023.

\bibitem[Podell et~al.(2023)Podell, English, Lacey, Blattmann, Dockhorn,
  M{\"u}ller, Penna, and Rombach]{podell2023sdxl}
Dustin Podell, Zion English, Kyle Lacey, Andreas Blattmann, Tim Dockhorn, Jonas
  M{\"u}ller, Joe Penna, and Robin Rombach.
\newblock Sdxl: Improving latent diffusion models for high-resolution image
  synthesis.
\newblock \emph{arXiv preprint arXiv:2307.01952}, 2023.

\bibitem[Qi et~al.(2023)Qi, Cun, Zhang, Lei, Wang, Shan, and
  Chen]{qi2023fatezero}
Chenyang Qi, Xiaodong Cun, Yong Zhang, Chenyang Lei, Xintao Wang, Ying Shan,
  and Qifeng Chen.
\newblock Fatezero: Fusing attentions for zero-shot text-based video editing.
\newblock In \emph{ICCV}, 2023.

\bibitem[Qin et~al.(2023)Qin, Zhang, Yu, Feng, Yang, Zhou, Wang, Niebles,
  Xiong, Savarese, et~al.]{qin2023unicontrol}
Can Qin, Shu Zhang, Ning Yu, Yihao Feng, Xinyi Yang, Yingbo Zhou, Huan Wang,
  Juan~Carlos Niebles, Caiming Xiong, Silvio Savarese, et~al.
\newblock Unicontrol: a unified diffusion model for controllable visual
  generation in the wild.
\newblock In \emph{NeurIPS}, 2023.

\bibitem[Rombach et~al.(2022)Rombach, Blattmann, Lorenz, Esser, and
  Ommer]{rombach2022high}
Robin Rombach, Andreas Blattmann, Dominik Lorenz, Patrick Esser, and Bj{\"o}rn
  Ommer.
\newblock High-resolution image synthesis with latent diffusion models.
\newblock In \emph{CVPR}, 2022.

\bibitem[Ronneberger et~al.(2015)Ronneberger, Fischer, and
  Brox]{ronneberger2015u}
Olaf Ronneberger, Philipp Fischer, and Thomas Brox.
\newblock U-net: Convolutional networks for biomedical image segmentation.
\newblock In \emph{MICCAI}, 2015.

\bibitem[Rout et~al.(2024)Rout, Chen, Ruiz, Caramanis, Shakkottai, and
  Chu]{rout2024semantic}
Litu Rout, Yujia Chen, Nataniel Ruiz, Constantine Caramanis, Sanjay Shakkottai,
  and Wen-Sheng Chu.
\newblock Semantic image inversion and editing using rectified stochastic
  differential equations.
\newblock \emph{arXiv preprint arXiv:2410.10792}, 2024.

\bibitem[Shin et~al.(2024)Shin, Kim, Lee, Lee, and Yoon]{shin2024edit}
Chaehun Shin, Heeseung Kim, Che~Hyun Lee, Sang-gil Lee, and Sungroh Yoon.
\newblock Edit-a-video: Single video editing with object-aware consistency.
\newblock In \emph{ACML}, 2024.

\bibitem[Singer et~al.(2024)Singer, Zohar, Kirstain, Sheynin, Polyak, Parikh,
  and Taigman]{singer2024video}
Uriel Singer, Amit Zohar, Yuval Kirstain, Shelly Sheynin, Adam Polyak, Devi
  Parikh, and Yaniv Taigman.
\newblock Video editing via factorized diffusion distillation.
\newblock In \emph{ECCV}, 2024.

\bibitem[Song et~al.(2024)Song, Cui, Zhang, Chen, Hong, and
  Jiang]{song2024doubly}
Xue Song, Jiequan Cui, Hanwang Zhang, Jingjing Chen, Richang Hong, and Yu-Gang
  Jiang.
\newblock Doubly abductive counterfactual inference for text-based image
  editing.
\newblock In \emph{CVPR}, 2024.

\bibitem[Team(2024)]{genmo2024mochi}
Genmo Team.
\newblock Mochi 1.
\newblock \url{https://github.com/genmoai/models}, 2024.

\bibitem[Tumanyan et~al.(2023)Tumanyan, Geyer, Bagon, and
  Dekel]{Tumanyan_2023_CVPR}
Narek Tumanyan, Michal Geyer, Shai Bagon, and Tali Dekel.
\newblock Plug-and-play diffusion features for text-driven image-to-image
  translation.
\newblock In \emph{CVPR}, 2023.

\bibitem[Wang et~al.(2024{\natexlab{a}})Wang, Pu, Qi, Guo, Ma, Huang, Chen, Li,
  and Shan]{wang2024taming}
Jiangshan Wang, Junfu Pu, Zhongang Qi, Jiayi Guo, Yue Ma, Nisha Huang, Yuxin
  Chen, Xiu Li, and Ying Shan.
\newblock Taming rectified flow for inversion and editing.
\newblock \emph{arXiv preprint arXiv:2411.04746}, 2024{\natexlab{a}}.

\bibitem[Wang et~al.(2024{\natexlab{b}})Wang, Bai, Wang, Qin, Chen, Li, Tang,
  and Hu]{wang2024instantid}
Qixun Wang, Xu Bai, Haofan Wang, Zekui Qin, Anthony Chen, Huaxia Li, Xu Tang,
  and Yao Hu.
\newblock Instantid: Zero-shot identity-preserving generation in seconds.
\newblock \emph{arXiv preprint arXiv:2401.07519}, 2024{\natexlab{b}}.

\bibitem[Wang et~al.(2023{\natexlab{a}})Wang, Jiang, Xie, Liu, Chen, Cao, Wang,
  and Shen]{wang2023zero}
Wen Wang, Yan Jiang, Kangyang Xie, Zide Liu, Hao Chen, Yue Cao, Xinlong Wang,
  and Chunhua Shen.
\newblock Zero-shot video editing using off-the-shelf image diffusion models.
\newblock \emph{arXiv preprint arXiv:2303.17599}, 2023{\natexlab{a}}.

\bibitem[Wang et~al.(2023{\natexlab{b}})Wang, Zhao, and
  Xing]{wang2023stylediffusion}
Zhizhong Wang, Lei Zhao, and Wei Xing.
\newblock Stylediffusion: Controllable disentangled style transfer via
  diffusion models.
\newblock In \emph{ICCV}, 2023{\natexlab{b}}.

\bibitem[Wu et~al.(2023)Wu, Ge, Wang, Lei, Gu, Shi, Hsu, Shan, Qie, and
  Shou]{wu2023tune}
Jay~Zhangjie Wu, Yixiao Ge, Xintao Wang, Stan~Weixian Lei, Yuchao Gu, Yufei
  Shi, Wynne Hsu, Ying Shan, Xiaohu Qie, and Mike~Zheng Shou.
\newblock Tune-a-video: One-shot tuning of image diffusion models for
  text-to-video generation.
\newblock In \emph{ICCV}, 2023.

\bibitem[Xie et~al.(2023)Xie, Zhang, Lin, Hinz, and Zhang]{xie2023smartbrush}
Shaoan Xie, Zhifei Zhang, Zhe Lin, Tobias Hinz, and Kun Zhang.
\newblock Smartbrush: Text and shape guided object inpainting with diffusion
  model.
\newblock In \emph{CVPR}, 2023.

\bibitem[Xu et~al.(2024)Xu, Tang, Cao, Zhang, Kong, Li, Deussen, and
  Lee]{xu2024headrouter}
Yu Xu, Fan Tang, Juan Cao, Yuxin Zhang, Xiaoyu Kong, Jintao Li, Oliver Deussen,
  and Tong-Yee Lee.
\newblock Headrouter: A training-free image editing framework for mm-dits by
  adaptively routing attention heads.
\newblock \emph{arXiv preprint arXiv:2411.15034}, 2024.

\bibitem[Yang et~al.(2023)Yang, Ding, Wang, Chen, Zhuang, and
  Shen]{yang2023object}
Zhen Yang, Ganggui Ding, Wen Wang, Hao Chen, Bohan Zhuang, and Chunhua Shen.
\newblock Object-aware inversion and reassembly for image editing.
\newblock \emph{arXiv preprint arXiv:2310.12149}, 2023.

\bibitem[Yang et~al.(2024)Yang, Teng, Zheng, Ding, Huang, Xu, Yang, Hong,
  Zhang, Feng, et~al.]{yang2024cogvideox}
Zhuoyi Yang, Jiayan Teng, Wendi Zheng, Ming Ding, Shiyu Huang, Jiazheng Xu,
  Yuanming Yang, Wenyi Hong, Xiaohan Zhang, Guanyu Feng, et~al.
\newblock Cogvideox: Text-to-video diffusion models with an expert transformer.
\newblock \emph{arXiv preprint arXiv:2408.06072}, 2024.

\bibitem[Ye(2024)]{ye2024ground}
Jong~Chul Ye.
\newblock Ground-a-video: Zero-shot grounded video editing using text-to-image
  diffusion models.
\newblock In \emph{ICLR}, 2024.

\bibitem[Zhang et~al.(2023)Zhang, Mo, Chen, Sun, and Su]{zhang2023magicbrush}
Kai Zhang, Lingbo Mo, Wenhu Chen, Huan Sun, and Yu Su.
\newblock Magicbrush: A manually annotated dataset for instruction-guided image
  editing.
\newblock \emph{NeurIPS}, 2023.

\bibitem[Zhang et~al.(2024)Zhang, Yang, Feng, Qin, Chen, Yu, Chen, Wang,
  Savarese, Ermon, et~al.]{zhang2024hive}
Shu Zhang, Xinyi Yang, Yihao Feng, Can Qin, Chia-Chih Chen, Ning Yu, Zeyuan
  Chen, Huan Wang, Silvio Savarese, Stefano Ermon, et~al.
\newblock Hive: Harnessing human feedback for instructional visual editing.
\newblock In \emph{CVPR}, 2024.

\bibitem[Zhao et~al.(2024)Zhao, Gu, Wu, Zhang, Liu, Wu, Keppo, and
  Shou]{zhao2024motiondirector}
Rui Zhao, Yuchao Gu, Jay~Zhangjie Wu, David~Junhao Zhang, Jia-Wei Liu, Weijia
  Wu, Jussi Keppo, and Mike~Zheng Shou.
\newblock Motiondirector: Motion customization of text-to-video diffusion
  models.
\newblock In \emph{ECCV}, 2024.

\bibitem[Zhao et~al.(2023)Zhao, Xie, Hong, Li, and Lee]{zhao2023make}
Yuyang Zhao, Enze Xie, Lanqing Hong, Zhenguo Li, and Gim~Hee Lee.
\newblock Make-a-protagonist: Generic video editing with an ensemble of
  experts.
\newblock \emph{arXiv preprint arXiv:2305.08850}, 2023.

\bibitem[Zuo et~al.(2025)Zuo, Zhang, Luo, Zhao, Zhang, Yang, and
  Wang]{zuo2025cut}
Zhichao Zuo, Zhao Zhang, Yan Luo, Yang Zhao, Haijun Zhang, Yi Yang, and Meng
  Wang.
\newblock Cut-and-paste: Subject-driven video editing with attention control.
\newblock \emph{Neural Networks}, 2025.

\end{thebibliography}
}

\clearpage
\setcounter{page}{1}
\maketitlesupplementary

\section{Pseudo Code}
\begin{figure}[t]
    \centering
    \includegraphics[width=1.0\linewidth]{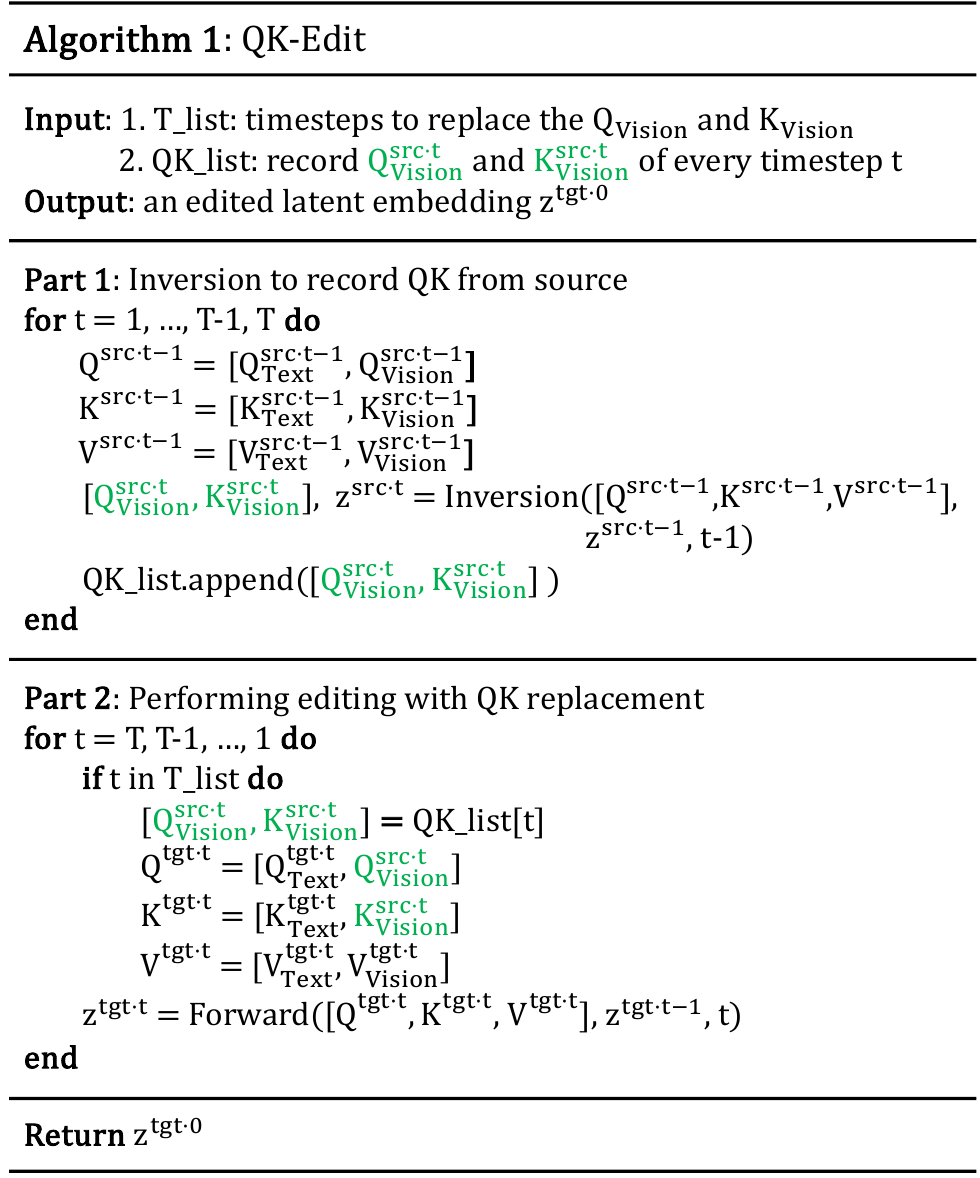}
    \caption{The pseudo code of QK-Edit.}
    \label{fig:supp_algo}
\end{figure}

To illustrate the simplicity of our QK-Edit, we provide the pytorch-style pseudo code for our algorithm in Fig.~\ref{fig:supp_algo}. This pseudo code demonstrates the straightforward implementation of QK-Edit.

\section{Additional Image Editing Results}
\begin{figure*}[t]
    \centering
    \vspace{-6mm}
    \includegraphics[width=0.88\linewidth]{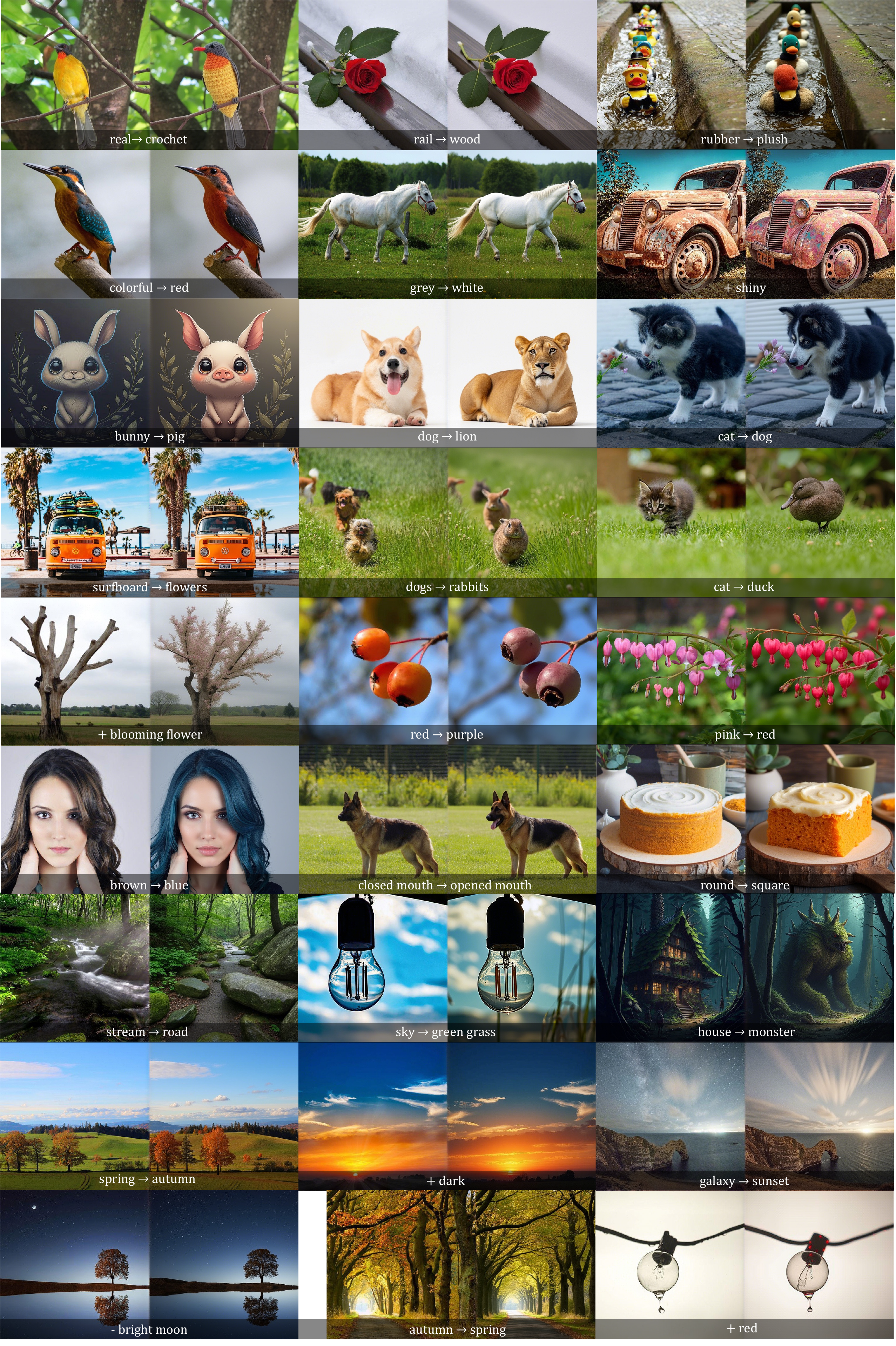}
    \vspace{-3mm}
    \caption{Additional image editing results.}
    \label{fig:supp_img}
\end{figure*}

Additional qualitative results for image editing are provided in Fig.~\ref{fig:supp_img}. These examples further demonstrate QK-Edit's ability to preserve fine-grained details, maintain structural consistency, and achieve precise semantic alignment across diverse editing scenarios.

\begin{figure}[h]
    \centering
    \includegraphics[width=1.0\linewidth]{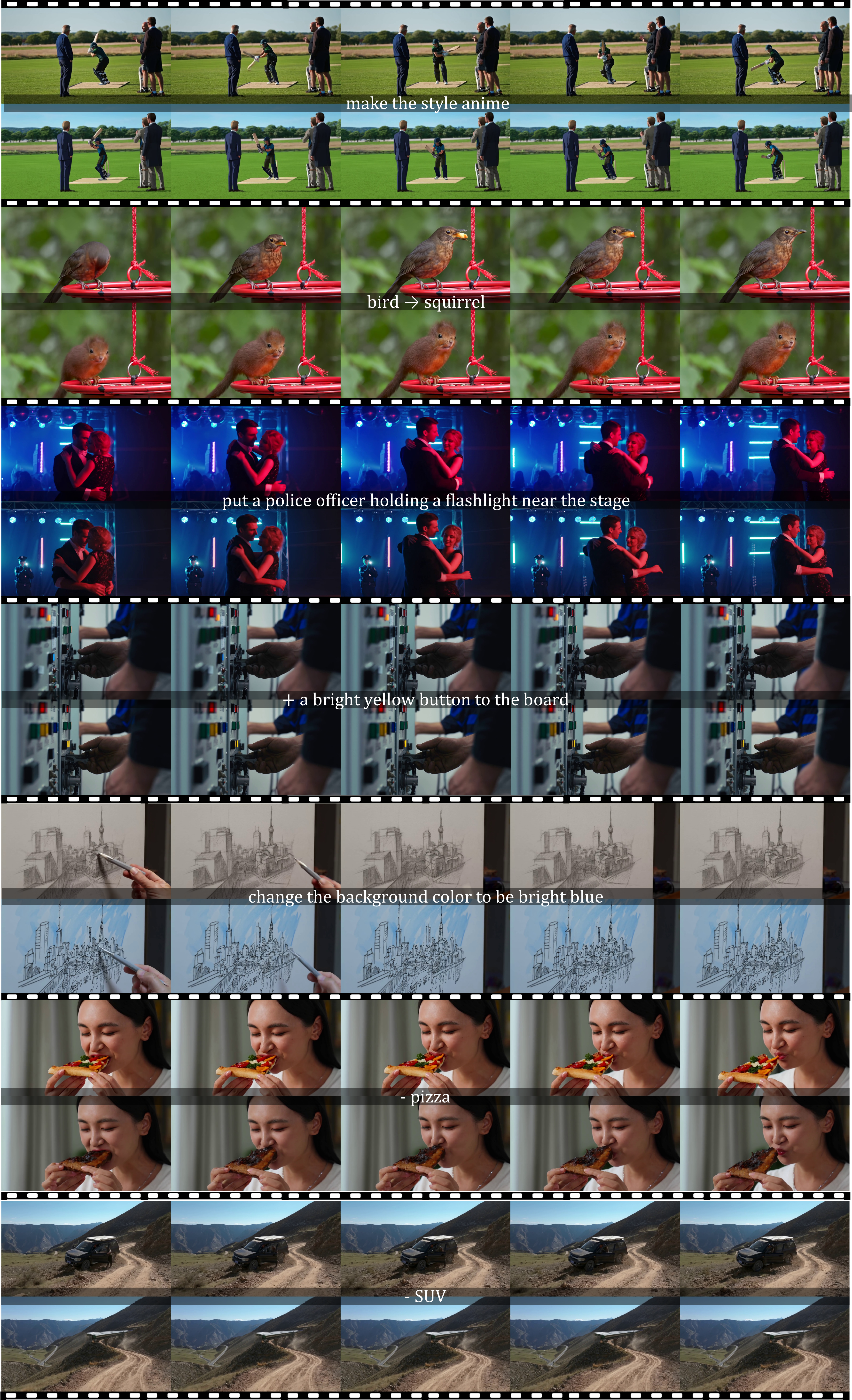}
    \vspace{-3mm}
    \caption{Failure cases and artifacts.}
    \label{fig:supp_artifacts}
\end{figure}

\section{Video Editing Results for High-Resolution}
Please refer to the accompanying MP4 files for video editing results at higher resolutions, longer frames, and two different aspect ratios. Specifically, the two files contain results for videos with dimensions of 1280×720 with 129 frames~($\texttt{1280\textunderscore 720\textunderscore 129.mp4}$) and 720×1280 with 65 frames~($\texttt{720\textunderscore1280\textunderscore65.mp4}$), showcasing the flexibility of our method on both landscape and portrait formats.

\section{Failure Cases and Artifacts}
As a training-free method, our performance is inherently bounded by the capabilities of the base model. We have discussed some failure cases in the Limitation section. More failure cases are illustrated in Fig.~\ref{fig:supp_artifacts}.


\end{document}